%% file: main_scai2024.tex
\documentclass[letterpaper, 10 pt, conference]{ieeeconf}  

\IEEEoverridecommandlockouts                              
\usepackage[utf8]{inputenc}
\usepackage[T1]{fontenc}

\usepackage{graphicx}
\usepackage{multirow} 
\usepackage{lipsum}
\usepackage{booktabs}
\usepackage{float} 
\usepackage{amsmath} 
\usepackage{stmaryrd} 

\usepackage[hidelinks]{hyperref} 

\title{\LARGE \bf
Enhancing Indoor Temperature Forecasting through Synthetic Data in Low-Data Environments}

\author{Zachari Thiry$^{1}$, Massimiliano Ruocco$^{1}$$^{2}$, Alessandro Nocente$^{1}$,  Michail Spitieris$^{1}$
\thanks{$^{1}$SINTEF AS, Strindvegen 4, 7034 Trondheim, Norway}
\thanks{$^{2}$Department of Computer Science, Norwegian University of Science and Technology, Sem Sælandsvei 9, 7034 Trondheim, Norway}
}

\begin{document}

\maketitle
\thispagestyle{empty}
\pagestyle{empty}

\begin{abstract}

\input{sections/abstract}

\end{abstract}

\section{Introduction}

\input{sections/introduction}

\section{State of the Art}

\input{sections/sota}

\section{Background theory}
\input{sections/background-theory}

\section{Methods}
\input{sections/methods}

\section{Results and Discussion}\label{sec:results}
\input{sections/results}

\section{Conclusions and Further Work}
\input{sections/conclusion}

\section*{Acknowledgement}
This research was carried out with the support of the ML4ITS project (312062) and RICO project (329730), funded by the Norwegian Research Council (NFR). We also thank Odne Andreas Oksavik for his technical support in data collection of the RICO dataset.

\addtolength{\textheight}{-12cm}   







\bibliographystyle{plain}
\bibliography{references}

\end{document}

%% file: sections/abstract.tex
Forecasting indoor temperatures is of paramount importance to achieve efficient control of HVAC systems. In this task, the limited data availability presents a challenge as most of the available data is acquired during standard operation where extreme scenarios and transitory regimes such as major temperature increases or decreases are \textit{de-facto} excluded. Acquisition of such data requires significant energy consumption and a dedicated facility, hindering the quantity and diversity of available data. To acquire such data, we make use of such a facility referred to as the Test-cell. Cost related constraints however do not allow for continuous year-around acquisition.
To address this, we investigate the efficacy of data augmentation techniques, particularly leveraging state-of-the-art AI-based methods for synthetic data generation. Inspired by practical and experimental motivations, we explore fusion strategies of real and synthetic data to improve forecasting models. This approach alleviates the need for continuously acquiring extensive time series data, especially in contexts involving repetitive heating and cooling cycles in buildings.
Our evaluation methodology for synthetic data synthesis involves a dual-focused approach: firstly, we assess the performance of synthetic data generators independently, particularly focusing on SoTA AI-based methods; secondly, we measure the utility of incorporating synthetically augmented data in a subsequent downstream tasks (forecasting).
In the forecasting tasks, we employ a simple model in two distinct scenarios: 1) we first examine an augmentation technique that combines real and synthetically generated data to expand the training dataset, 2) Second, we delve into utilizing synthetic data to tackle dataset imbalances.
Our results highlight the potential of synthetic data augmentation in enhancing forecasting accuracy while mitigating training variance. Through empirical experiments, we show significant improvements achievable by integrating synthetic data, thereby paving the way for more robust forecasting models in low-data regime.

%% file: sections/introduction.tex
Indoor temperature forecasting predicts future temperature values in the different rooms of a building, leveraging historical data and environmental factors for proactive Heating, Ventilation and Air Conditioning (HVAC) system management and comfort optimization. 
The European Union emphasizes the importance of efficient building energy management systems to achieve sustainability goals, given that buildings contribute to 40\% of energy consumption and 36\% of CO\textsubscript{2} emissions in the EU \cite{european2010directive}. HVAC systems, responsible for the majority of energy consumption in buildings, significantly influence both household comfort and environmental impact.

Typically, heating and cooling systems in buildings are controlled by a schedule. 
This type of regulation does not take into account factors such as outdoor weather, solar radiation, and changes in occupancy, and therefore can lead to excessively heating (or cooling) of a room thus creating discomfort for the occupants.
A solution that, instead, makes use of the predicted room temperature as an input, can lead to a better comfort for the occupants while achieving consistent savings in energy use.

Machine learning models have demonstrated superiority over traditional physics-based methods in indoor temperature forecasting \cite{fi13100242, en11020395, BELLAGARDA2022104314}. Currently, Recurrent Neural Networks (RNNs), such as Long Short-Term Memory (LSTM) networks, remain a practical choice for such forecasting tasks \cite{DEIHIMI2012327,Alawadi2022,FANG2021111053}. Our approach differs from those aforementioned as we do not focus on finding the best forecaster. Instead, we seek to enhance forecasting as a whole, in particular in low-data environments.

Synthetic data generation is a rapidly growing field \cite{mumuni2024survey, torres2021deep, Iglesias_2023, vogt2022synthetic, fi13100242, Zhou2020} within the realm of data augmentation, predominantly relying on variations of Generative Adversarial Networks (GANs). 
Its applications span across diverse domains, from medicine to maintenance tasks \cite{math9182336, 8363576, fatigueZhai2022, kimInfrared2022}. However, the impact of synthetic data augmentation on temperature forecasting in low-data environments remains relatively unexplored. This study endeavors to address this gap by augmenting forecasters with synthetic data and evaluating their performance in subsequent tasks. Through this investigation, we aim to uncover the key effects and potential benefits of synthetic data augmentation in enhancing temperature forecasting accuracy amidst data scarcity.

\textbf{Roadmap:} We begin by reviewing the state-of-the-art and existing synthesizers. Next, we look into our methodology, including the fusion of synthetic and real samples and addressing class imbalance. Finally, we present our experimental results.

%% file: sections/sota.tex
The literature surveys, as evidenced by \cite{Iglesias_2023}, \cite{mumuni2024survey}, and \cite{torres2021deep}, commonly delineate modern approaches to time series data augmentation into three broad categories: traditional methods, GAN-based techniques, and Auto-Encoder-based techniques. Traditional methods such as homogeneous scaling and rotation are noted for their cost-effectiveness and simplicity. However, they often disrupt temporal relationships within the time series. Consequently, contemporary approaches lean towards generative models to better preserve temporal dynamics.

For instance, \cite{Iglesias_2023} discusses 23 methods, including 9 Variational Auto-Encoder based (VAE) and 14 GAN-based techniques. Despite the widespread use of GANs, they are prone to convergence issues, particularly in low-data contexts, although they yield more diverse data. The over representation of GANs in the generic synthetic generation literature is furthermore confirmed by the reports from \cite{mumuni2024survey}.

Synthetic data augmentation has demonstrated its effectiveness in enhancing forecasting performance for time series data, spanning various domains. For example, \cite{vogt2022synthetic} showcases successful synthetic data generation in the context of renewable power plant energy forecasting, leveraging physical models and generic weather prediction systems.

In addition, Machine Learning approaches have been explored extensively in this domain. Notably, \cite{Zhou2020} successfully predicts emerging technologies within a year by augmenting synthetic patent data using GANs. Then, \cite{9670649} proposes a traditional ML method, K-means, for synthetic data generation, albeit with mixed results in forecasting methods involving deep learning architectures like LSTMs.

In summary, the literature offers a variety of methods for time series data augmentation, ranging from traditional to advanced machine learning techniques. Given the diverse landscape of options, we opt to focus on deep learning methods for data synthesis in our work. Specifically, we choose to study GANs for their ability to generate diverse data and VAEs for their ease of training, especially in low-data scenarios. We highlight three notable models from the literature \cite{Yoon2019, DoppleGANger, lee2023vector}, and detail their mechanisms in the subsequent section.

%% file: sections/background-theory.tex
\subsection{TimeGAN}
Initially proposed in \cite{Yoon2019}, it represents a pioneering effort in exploring the capabilities of generative adversarial network architectures for time series data. It incorporates serveral strategies to enhance efficiency:
firstly, it employs both an adversarial loss and a supervised loss, combining the control provided by supervised learning with the flexibility inherent in unsupervised GAN models.
Additionally, TimeGAN utilizes a dimension reduction technique involving embedding and recovery networks. These networks map features to latent representations, effectively reducing the dimensionality of the time series data. This approach capitalizes on the fact that the temporal dynamics of time series data can often be captured in a lower-dimensional space relative to the length of the series, thereby simplifying the tasks performed by the GAN.
Furthermore, the generator and discriminator operate within the latent space. TimeGAN adopts a joint training approach for the embedding and the generative network. This strategy facilitates the learning of temporal relationships by the generator.


\subsection{DoppleGANger}

Developed as a versatile network-time series synthesizer, DoppleGANger was designed to address fidelity problems between measurements and their associated data, to better capture long-term correlations within time series data, and to mitigate issues such as mode collapse in generative models. 

To address fidelity concerns, DoppleGANger introduces an auxiliary discriminator dedicated to metadata generation ; although this aspect is not utilized in the context of this article. To mitigate mode collapse, it implements a strategy that constrains generation to randomized min-max values at each iteration, which are later scaled back to realistic ranges. This technique ensures diversity in generated samples and effectively combats mode collapse.
Moreover, DoppleGANger modifies the canonical GAN framework by integrating LSTM cells to better capture temporal dependencies. To mitigate memory loss associated with RNN cells, DoppleGANger introduces the concept of \textit{batched generation}, enabling the simultaneous generation of multiple records at each cell pass instead of the traditional single-step generation approach. This enhancement significantly improves the efficiency and memorisation effectiveness of the generative process.


\subsection{TimeVQVAE}

TimeVQVAE pioneers the application of Vector Quantization techniques to tackle the time series generation challenge, introducing several novel features: firstly, they employ Vector Quantization of the latent space with VQVAE \cite{lee2023vector}, a type of variational Auto-Encoder that leverages vector quantization to discretize the latent space while learning the prior distribution. This approach ensures that VQ-VAE avoids posterior collapse by learning a quantized latent space instead of constraining it, for example, by a Gaussian distribution.

Moreover, TimeVQVAE adopts a modified MaskGIT \cite{chang2022maskgit} prior learning process for the sampling phase, which is asserted to not only accelerate the process but also enhance the quality and diversity of generated samples.

Additionally, TimeVQVAE operates on a modified space: initially, time series data is shifted to a time-frequency space using Discrete Fourier Transform. Subsequently, separate sets of VQ-VAEs are trained for both low-frequency and high-frequency generation tasks, not only facilitating the learning process, but also ensuring the preservation of key features in both components.


%% file: sections/methods.tex
\subsection{Methods for data acquisition and processing}

\subsubsection{Description of the dataset}
The data in question has been acquired over a dedicated test facility\footnote{Link to the laboratory used for acquisition: \url{https://www.sintef.no/en/all-laboratories/zeb-test-cell-laboratory/}} and is stored under a tabular time series format of size ($N=59,040$, $D=81$), acquired at a rate of $1 \text{ min}^{-1}$ and following the principles detailed in the section below.

\begin{figure*}
    \centering
    \includegraphics[width=1\linewidth]{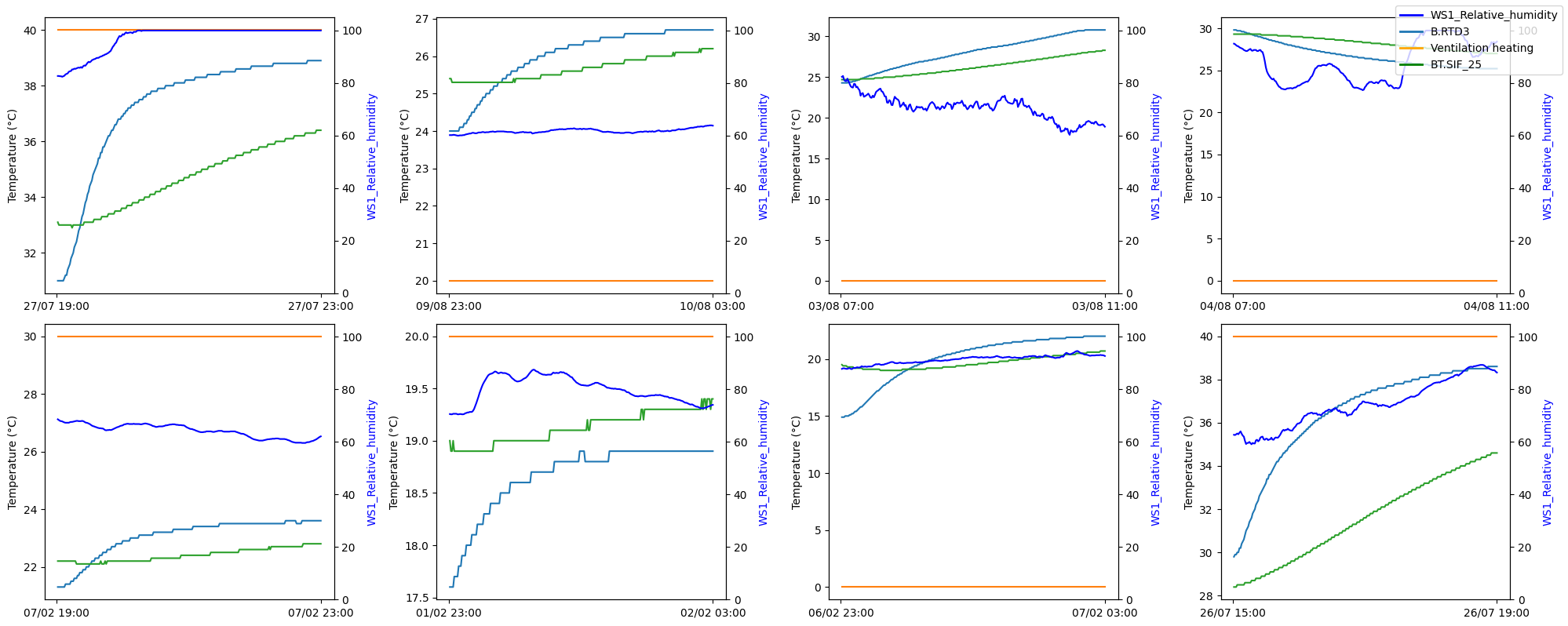}
    \caption{Visual observations from the dataset: four dimensions from eight random series.}
    \label{fig:multivariate-samples}
\end{figure*}

Regarding the dimensions, we define a series as a vector of shape $(240,D)$, constituting a sequence of 240 consecutive rows from the dataset, starting from the beginning. In essence, each series is D-dimensional, encapsulates four hours of data acquisition, and it is uniquely characterized by a phase and a step. See Figure \ref{fig:multivariate-samples} for visual examples of a series.

The dataset is acquired over four distinct phases, each yielding a subset denoted as RICO<X>, where X represents the acquisition number.

\begin{itemize}
    \item RICO1: collected between July and August 2023 spanning over 17 days, encompassing 102 series, equivalent to 24,480 rows. RICO1 exhibits some inconsistencies due to sub-optimal tuning of the acquisition facility. Although flawed, most data points exhibit "normal" behaviour. 
    
    \item RICO2: collected in October 2023 spanning over 10 days, encompassing 60 series, equivalent to 14,400 rows. In this acquisition, we introduced one hour of downtime, or \textit{"free fall"}, at the end of each four-hour point recording: in essence, the constraints are stopped and 'natural' heat exchanges are the only ones that remain.

    \item  RICO3: collected in January 2024, spanning over 4 days, encompassing 24 series, equivalent to 5,760 rows. RICO3 was acquired with sequences of sixteen hours of constraint for each series instead of four, with its last four hours left as free fall. To ensure compatibility with other datasets, only the first four hours of each series will be utilized, for a total of 6 useful series.

    \item RICO4: collected in February 2024, spanning over 10 days, encompassing 60 series, equivalent to 14,400 rows. This acquisition followed a similar protocol to RICO1, featuring fixed actuators tuning and no free fall. As of today, RICO4 stands as the phase that produced the highest quality samples.
\end{itemize}

The features of the dataset can be divided into five categories:

\begin{itemize}
\item \textbf{Identifiers}: These include Phase, Step, and Flag. Phase and Step serve to uniquely identify a point, while Flag enables the flagging of points we opt not to utilize.
\item \textbf{Setpoints}: These variables (EC3, SB43, B46 and SB47) represent the setpoints of the four HVAC systems within the test cell (two heaters and two coolers) that we can control. They are randomly adjusted every four hours to introduce diversity into the dataset.
\item \textbf{Features of Interest}: These entail variables like internal air temperature, which constitute the focal points of our predictive experiments.
\item \textbf{Environmental Variables}: This category encompasses external temperatures, wind direction, sun radiation, dew point, and other weather related metrics.
\item \textbf{Control Features}: Examples include JP40\_head, and pid.EC3.enabled. They primarily serve to verify data integrity but hold marginal relevance in machine learning contexts.
\end{itemize}

In each phase, the acquisition process unfolds as follows: a random combination of set points is generated based on predefined permissible values. These values reflect the temperatures reached by the HVAC actuators themselves and not those reached by the room temperature. For example, for heaters, these values may include [off, 20°C, 40°C, 60°C]. Every four hours, this combination of set points is dispatched to various system actuators. Each unique combination generates a single series.

\subsubsection{Feeding to the models}\label{ssec:model-feeding}

All acquired series are utilized for our analysis; however, certain points are manually excluded based on specific criteria:

\begin{itemize}
    \item Series from RICO1 exhibiting unexpected behaviors are identified and excluded from the analysis, totaling 19 series.
    
    \item Series from RICO2, acquired over a duration of (3h constrained + 1h free fall), are excluded as they differed from the intended format of 4h constrained.
    
    \item A subset of series from RICO3 (only 6 out of the total 24 series) is utilized, while others are excluded from the analysis.
    
    \item Series from RICO4 with missing values are identified and excluded from the analysis.
\end{itemize}

Each series earmarked for exclusion from our analysis is labeled with a tag point, denoted as 1 for inclusion and 0 for exclusion.

Regarding transformations, we first apply standard scaling to the data. Subsequently, the data is restructured from shape (N,D) to (N,L,C), where L represents the sequence length (240 here), and C represents the channels. In our case, C equals 1, focusing solely on one dimension (B.RTD3), which represents the temperature at the center of the room.

Ultimately, the dataset is partitioned into distinct training and testing sets. A fraction of 0.2 of the series from each phase is reserved for the test set, with the remaining series allocated to the training set. Consequently, the training set \textit{train\_real} comprises 116 series, while the testing set \textit{test\_real} contains 31 series.

\subsubsection{Data labeling methodology}

To label our dataset systematically, we follow these steps:

\begin{enumerate}
    \item \textbf{Subset Selection}: We focus on the initial 3 hours of each series to capture relevant data, excluding the last hour, which often represents a stable regime distinct from the rest of the data.
  
  \item \textbf{Smoothing}: Applying a 5-large moving average with edge-repetition padding to smooth the data, reducing noise.
  
  \item \textbf{Derivative Calculation}: Local derivatives are computed to capture trends in value changes.
  
  \item \textbf{Label Classes}:
    \begin{itemize}
      \item \textbf{Monotonic Positive (0)}: Showcases consistent increase in trends.
      \item \textbf{Monotonic Negative (1)}: Showcases consistent decrease in trend.
      \item \textbf{Non-Monotonic (2)}: Exhibits fluctuations or irregularities in trend.
    \end{itemize}
\end{enumerate}

These classes enable our synthesizers to effectively capture diverse trends present in the data.

\subsection{Comparing the synthetisers}

\subsubsection{Methods for evaluating the synthesizers}
In this study, we evaluate the performance of TimeVQVAE under various scenarios. We compare its results and synthetic capabilities with two other synthesizers: TimeGAN, serving as a baseline despite expected limitations, and the DoppleGANger model.

Our code base builds upon the implementations from \cite{lee2023vector} for TimeVQVAE, the gretel-ai python library for DoppleGANger, and J.Yoon's official implementation of TimeGAN \cite{Yoon2019}.

We initially train on 116 series extracted from all  RICO phases as detailed in \ref{ssec:model-feeding}. Training specifics for each model are as follows:
\begin{itemize}
    \item TimeGAN: Various settings were experimented with, but no satisfactory results were achieved.
    \item DoppleGANger: Sequence length of 240, batch size 8, and 1000 epochs.
    \item TimeVQVAE: After manual hyper-parameter tuning, the base parameters proved optimal: 2000 epochs for training the VQVAE and 10000 epochs for prior learning.
\end{itemize}

In evaluating the synthesizers, we employ a mix of traditional metrics alongside a utility metric which, in downstream tasks inspired by \cite{DoppleGANger} and \cite{Yoon2019}. The traditional algorithms include:

\begin{itemize}
    \item t-SNE (t-stochastic Neighborhood Embedding):  is a technique for dimensionality reduction, particularly effective for visualizing high-dimensional datasets in lower-dimensional spaces. It captures the local structure of the data, offering insights into its dependencies and feature relationships.
    \item PCA (Principal Component Analysis): identifies the principal components of a dataset, reducing its dimension while retaining as much variance as possible. It aids in understanding the underlying structure and dominant patterns within the data.
\end{itemize}

In addition to these traditional metrics, we will also rely on visual observation of the samples to complement the quantitative analyses. This holistic approach ensures a comprehensive evaluation of the synthesizers' performance.

Lastly, our incorporation of the utility metric, provides an alternative perspective on the synthesizers' performance, highlighting their practical utility in forecasting tasks. The next chapter dives into the specifics of this metric and its implications for our study.

\subsection{Forecasting Utility}\label{sec:forecasting_utility}
\subsubsection{Utility metric}
This utility metric enables us to gauge the effectiveness of our synthesized samples in real-world forecasting tasks. Specifically, we conduct controlled experiments where a simple forecasting model is trained on a baseline dataset. Subsequently, we reduce or augment the dataset with synthetic samples and analyze the outcomes of these experiments.

In our study, we employ a straightforward one-layer LSTM model followed by a fully connected layer as our forecasting model. The hyper parameters of this model are tuned manually using a training set identical to the remainder of our experiments. Predictions are made for the subsequent thirty minutes, with the data being sub-sampled by a factor of 10 before being fed into the model.

\subsubsection{Experiment: General data augmentation}
\label{sssec:general-augmentation}
In our first experiment, we start by training an instance of the chosen synthesizer on \textit{train\_real}. From the synthesizer, we sample 256 points denoted \textit{synth} from the trained synthesizer. We utilise \textit{synth} to construct training sets for the three following strategies: 
\begin{itemize}
    \item \textbf{TRTR:} or "Train Real, Test Real" where the train set consists of \textit{train\_real} and the test set consists in \textit{test\_real}. This corresponds to a normal control experiment.
    \item \textbf{TSTR:} or "Train Synthetic, Test Real" where the train set consists of \textit{synth} and the test set consists in \textit{test\_real}.
    \item \textbf{TRSTR:} or "Train Real and Synthetic, Test Real" where the train set consists in both \textit{train\_real} and \textit{synth}, and the test set consists in \textit{test\_real}.
\end{itemize}

We will employ a TRTR strategy as the baseline approach to establish a reference point for performance evaluation. Additionally, we will explore the alternative strategies TRSTR and TSTR, which involve the integration of synthetic data. By comparing the performance across these strategies, we aim to assess the effectiveness of incorporating synthetic data in our forecasting models.

\subsubsection{Experiment: Class imbalance}
In our second experiment, we tackle the question of class imbalance. Class imbalance is a typical Machine Learning problem where if one class in under-represented within the training set, the model's performance can be hindered in deployment for points belonging to that class. 
To see if synthetic data points can solve this issue, we experiment by artificially under-sampling a class from the training set and after, leveraging the conditional generation capabilities of our synthesizer to over-sample from this class, thus restoring balance in the forecaster training set.

We construct imbalanced training sets as a subset of our main training set: let $i$ be the class index, $n_i^{init}$ represent the number of samples initially within that class, and $n_i^{ablated}$ denote the number of samples remaining after ablation. We define the ablation ratio $r$ as:

\[
r = \frac{n_i^{ablated}}{n_i^{init}}
\]
We then create a new training set denoted $Set_{i,r}$ where we remove $n_{missing} = n_i^{init} - n_i^{ablated}$ samples from class $i$. Initially, we train a series of synthesizers on the $Set_{0,r} \text{ where } r \in R = \{0.25, 0.5, 0.75, 1.0\}$. This process is repeated for all three classes, resulting in the training of 12 synthesizers, each tailored to a specific type of training data, and denoted $\Sigma_{i,r}$.

Subsequently, for each of these 12 scenarios, we conduct the following training procedures:

\begin{itemize}
    \item We train a set of 'baseline' LSTMs on $Set_{i,r}$
    \item We train a set of 'test' LSTMs on $Set_{i,r}$ to which are appended a unique set of $n_{missing}$ points samples from $\Sigma_{i,r}$ for each training instance.
\end{itemize}
Results are discussed in the Results section \ref{sec:results}.

\subsection{Metrics used}
Our experiments evaluation will be based on four different standard metrics for time series forecasting 
\cite{torres2021deep}: 
\begin{itemize}
    \item MSE is a standard metric for measuring losses in continuous regression problems. It measures the average of the squared differences between forecasted and actual values, emphasizing large errors due to squaring terms. While it provides insight into the average squared deviation, MSE is sensitive to outliers and offers limited interpretability.
        \begin{equation}
        MSE = \frac{1}{n} \sum_{t=1}^{n} (y_t - \hat{y}_t)^2
        \end{equation}
    \item MAE computes the average of the absolute differences between forecasted and actual values. It is simple, easy to interpret, has symmetric penalisation and is robust to outliers. However, MAE does not emphasize large errors as much as MSE does.
        \begin{equation}
        MAE = \frac{1}{n} \sum_{t=1}^{n} |y_t - \hat{y}_t|
        \end{equation}
    \item MAPE calculates the average of the absolute percentage errors between actual and forecasted values. It offers easy interpretation in percentage terms and reflects the relative error size. However, closer to 0 and due the the non linearity of the inverse function, MAPE can be either undefined or easily influenced by outliers. ($y_t$ close to zero and $\hat{y}_t \neq y-t$)
    \begin{equation}
    MAPE = \frac{1}{n} \sum_{t=1}^{n} \left| \frac{y_t - \hat{y}_t}{y_t} \right|
    \end{equation}
    \item MASE provides a standardized measure of forecast accuracy by comparing a model's performance to that of a naïve forecast. Its calculation involves the mean of the absolute errors divided by the mean absolute error of a naïve forecast. MASE is scale independent, symmetric with respect to over and under predictions, but depends on a naive forecaster and is thus different for every dataset encountered.

    In most cases, $n$ is chosen as the seasonality. However, due to the absence of seasonality in our data, we opted to use n as the forecasting output window length.
    \begin{equation}
    MASE = \frac{1}{n} \sum_{t=1}^{n} \frac{|y_t - \hat{y}_t|}{\frac{1}{n-1} \sum_{t=2}^{n} |y_t - y_{t-1}|}
    \end{equation}
    \label{eq:std-mases}
\end{itemize}

%% file: sections/results.tex
\subsection{Synthetiser performance}

\begin{figure*}
    \centering
    \includegraphics[width=1\linewidth]{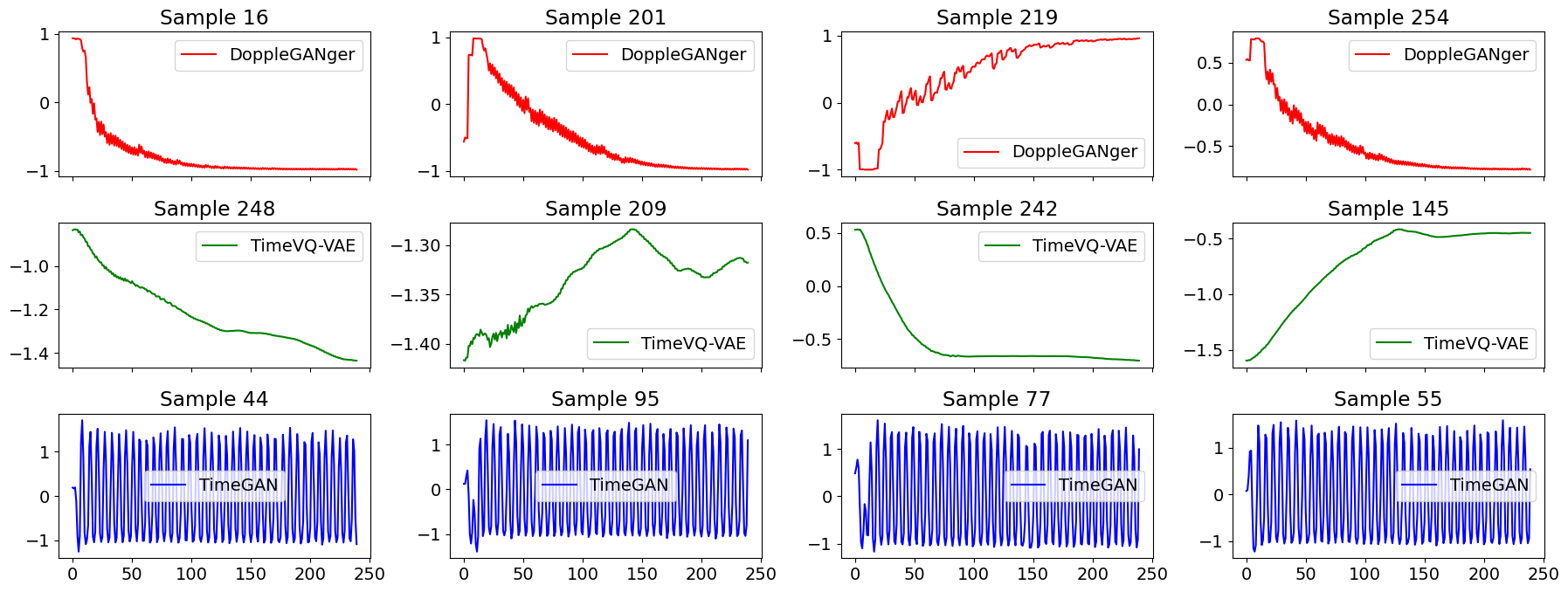}
    \caption{Example sample series from the trained synthetizers. From top to bottom: DoppleGANger, TimeVQVAE, TimeGAN.}
    \label{fig:synthetiser-samples}
\end{figure*}

The training of our forecasters show the following results:
Upon visual inspection of the generated samples in Figure \ref{fig:synthetiser-samples}, it appears clearly, and despite our best efforts, that TimeGAN (last row) was not able to converge and is thus disconsidered from the downstream tasks.
Concerning the DoppleGANger model (first row), it demonstrates an ability to capture the general trend from the training set, but introduces a high frequency parasite component, which resembles samples with high noise to signal ratio (see for example, samples 201 and 254).
Lastly, TimeVQVAE appears to show some more diversity than its counterpart, and is free of the noisy artifacts present in DoppleGANger.

\begin{figure}[H]
    \centering
    \includegraphics[width=1\linewidth]{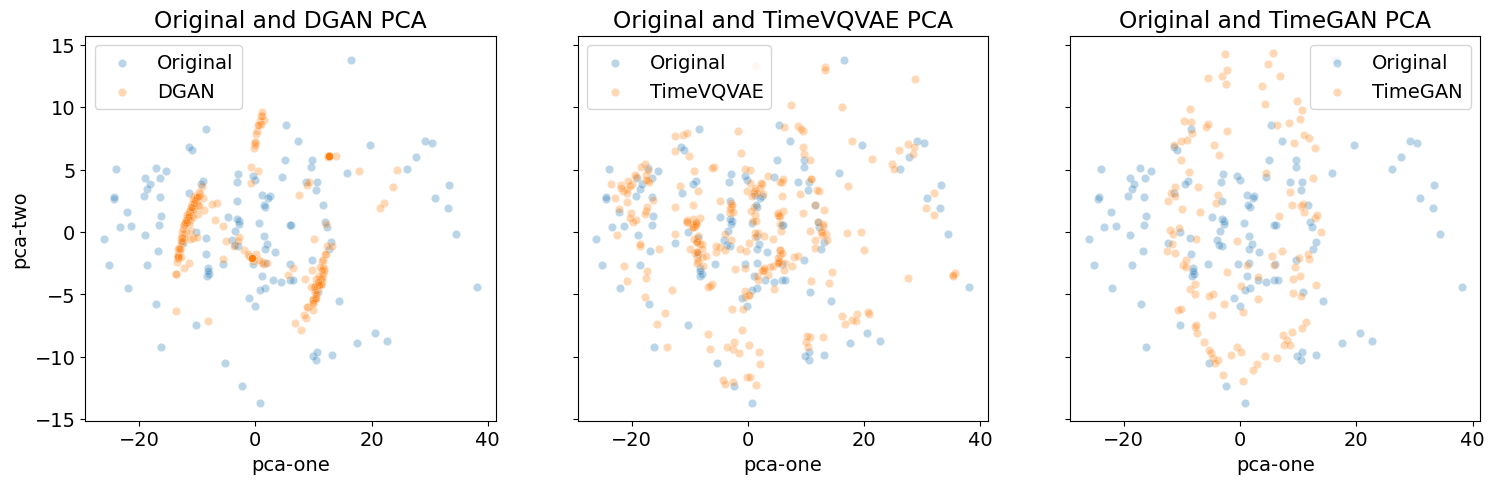}
    \caption{PCA Analysis (first two components) of the generated samples from all synthesizers, compared to the training samples.}
    \label{fig:synthetiser-pca}
\end{figure}

In Figure \ref{fig:synthetiser-pca}, which presents the PCA results from all synthesizers, one can observe the diversity of some unconditionally generated samples. Disregarding results from TimeGAN, the following observations emerge:

\begin{itemize}
\item DoppleGANger (left) is capable of generating plausible samples; however, it seems to struggle with generalization, as it only generates from limited regions of the underlying data distribution.
\item Conversely, TimeVQVAE (center) not only produces plausible samples but also appears to cover the entirety of the data space in the two-dimensional PCA mapping.
\end{itemize}

\begin{figure}[ht]
    \centering
    \includegraphics[width=1\linewidth]{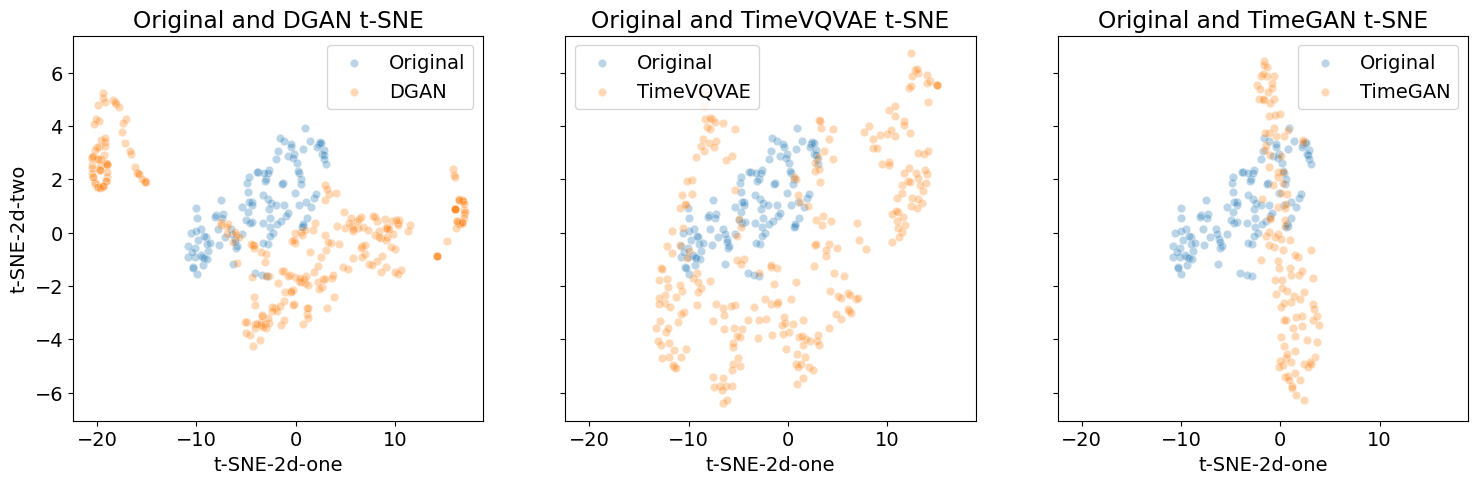}
    \caption{t-SNE Anslysis (first two components) of the generated samples from all synthesizers, compared to the training samples.}
    \label{fig:synthetiser-tsnes}
\end{figure}

These observations are furthermore confirmed upon observation of the t-SNE of the generated samples shown in Figure \ref{fig:synthetiser-tsnes}. While none of the studied models exhibit perfect overlap between image and data space, we observe that DoppleGANger exhibits correlations groups scattered apart from each other nonexistent in the real data space, and TimeVQVAE, though there is no scattering, appears to extend beyond the boundaries set by the original data t-SNE. It is however challenging to analyse the distances in t-SNE since they do not necessarily reflect actual distances between points \cite{wattenberg2016how}.

\subsection{Experimental study 1: synthetic augmentation}

We conduct an analysis comparing the effectiveness of the three strategies outlined in Section \ref{sssec:general-augmentation}. Initially, we train 100 forecasters using a TRTR strategy (\ref{sec:forecasting_utility}), on \textit{train\_real}. Subsequently, we train: 
\begin{itemize}
    \item 100 forecasters using a TSTR strategy, where each forecaster is given a unique set of synthetic series generated by the synthesizer
    \item 100 forecasters using TRSTS strategy, where each forecaster is also given a new set of series.
\end{itemize}

\begin{figure}[ht]
    \centering
    \includegraphics[width=1\linewidth]{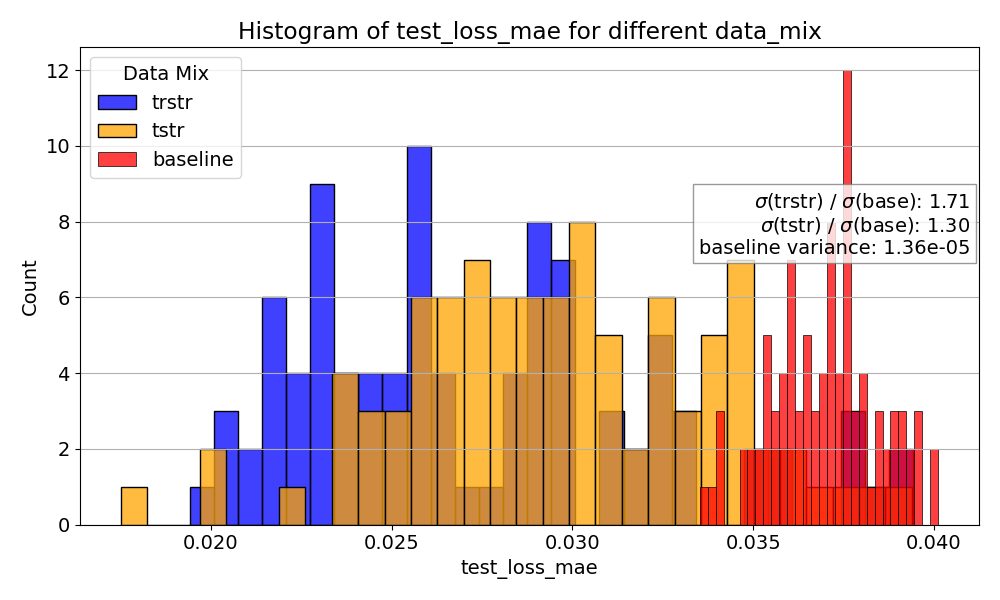}
    \caption{Histogram: MAE loss - forecast over 100 iterations for different strategies, excluding the outmost 5\% outliers.}
    \label{fig:mae}
\end{figure}

\begin{figure}[ht]
    \centering
    \includegraphics[width=1\linewidth]{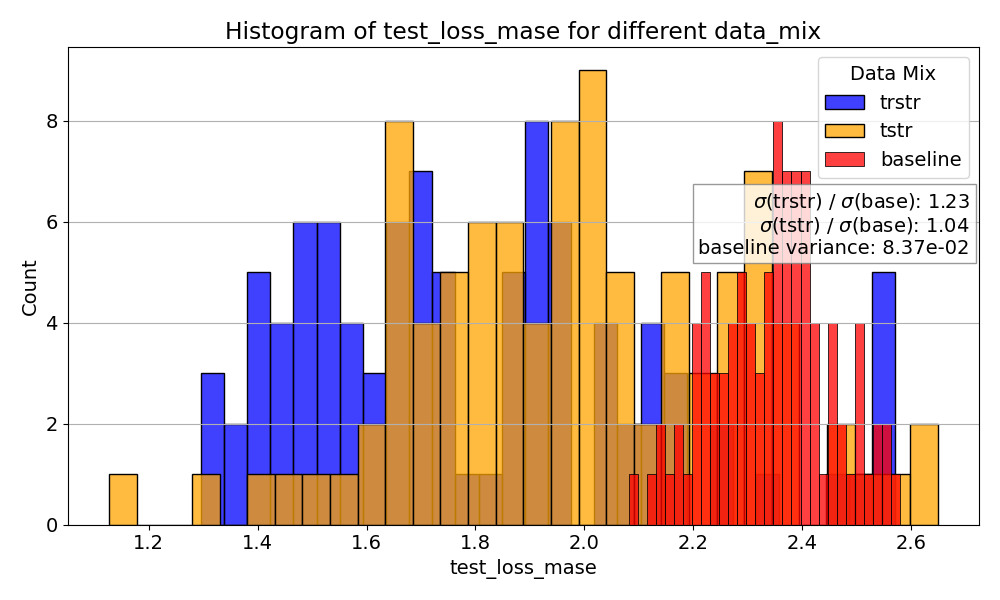}
    \caption{Histogram: MASE loss - forecast over 100 iterations for different strategies, excluding the outmost 5\% outliers.}
    \label{fig:mase}
\end{figure}

Figures \ref{fig:mae} and \ref{fig:mase} present histograms of the MAE test losses (lower is better) where we observe the following: overall performance improves on average with increased training data samples. These are the cases where synthetically generated samples have been introduced in the training process. Specifically, a significant performance improvement is observed with the use of only synthetic samples for training (TSTR scenario), and further improved when combining both synthetic and real samples (TRSTR) for training.

\input{artifacts/exp1_table_mu}
\input{artifacts/exp2_table_var}
We observe the following key points:
\begin{itemize}
    \item Table \ref{table-mean-exp1} demonstrates an overall performance improvement with increased data volumes. Significant enhancement is observed with the addition of synthetic samples (TSTR scenario), further augmented when combining both sythetic and real samples (TRSTR).
    \item In Table \ref{table-var-exp1}, the variance remains consistent across all strategies, with a general trend favouring the baseline. We believe that this increase in variance is due to the inherent variability introduced by the inclusion of newly generated samples in the TRSTR and TSTR strategies.
\end{itemize}
The behaviour observed on the MAE test losses is consistent across all four metrics employed.

\subsection{Experimental study 2: Class balancing}\label{ssec:exp2}
\begin{figure}[ht]
    \centering
    \includegraphics[width=1\linewidth]{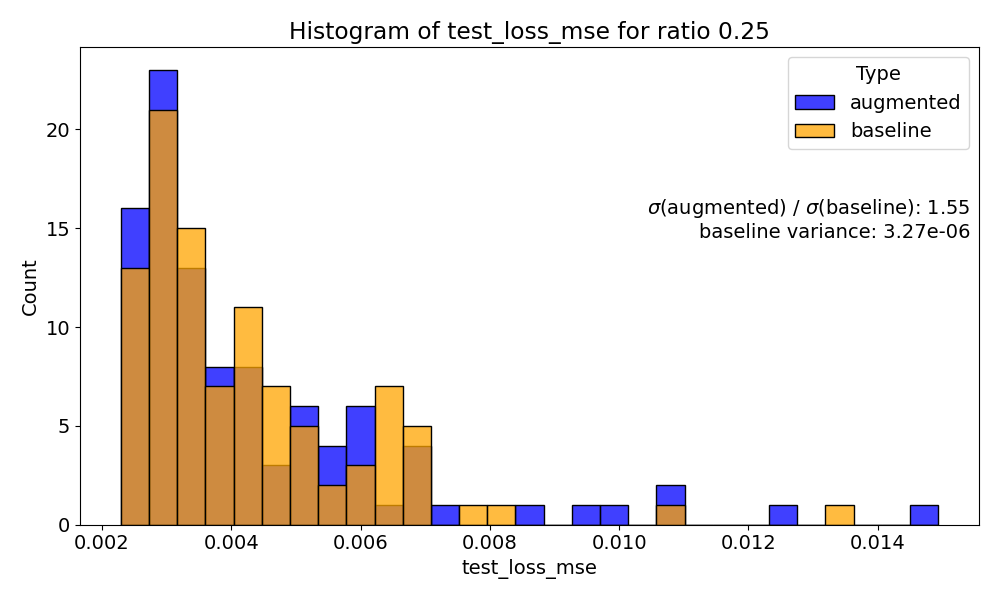}
    \caption{Histogram: Analysis of imbalanced vs augmented forecasting - Ablation ratio of .25 - Results for 100 runs.}
    \label{fig:dist_analysis_exp2_r0.25}
\end{figure}

In this experiment, we conduct another analysis to evaluate the effectiveness of augmenting imbalanced datasets with synthetic samples for prediction tasks.
Figure \ref{fig:dist_analysis_exp2_r0.25} illustrates one such scenario, where we selectively remove 75\% of the samples from class 0. 

Tables \ref{table-mean-exp2} and \ref{table-var-exp2} present aggregated results for the mean and variance, respectively, across 100 runs of the test metrics.
Interestingly, our analysis reveals no significant discernible improvement or deterioration in performance, as evidenced by the overlapping likelihood distributions of both the baseline and the augmented scenarios. However, regarding variance, our observations vary depending on the metric utilized and the ratio, with fluctuations ranging from a minor 0.14\% decrease to a more substantial 55\% increase. We hypothesise that such an increase might be caused by the nature of the datasets which remain strictly identical in all the baseline runs but is unique to each test run.

Despite these insights, our experiments alone do not offer conclusive explanations for the observed behavior. Further investigation is necessary to fully understand the underlying mechanisms driving these findings.

\begin{figure}[ht]
    \centering
    \includegraphics[width=1\linewidth]{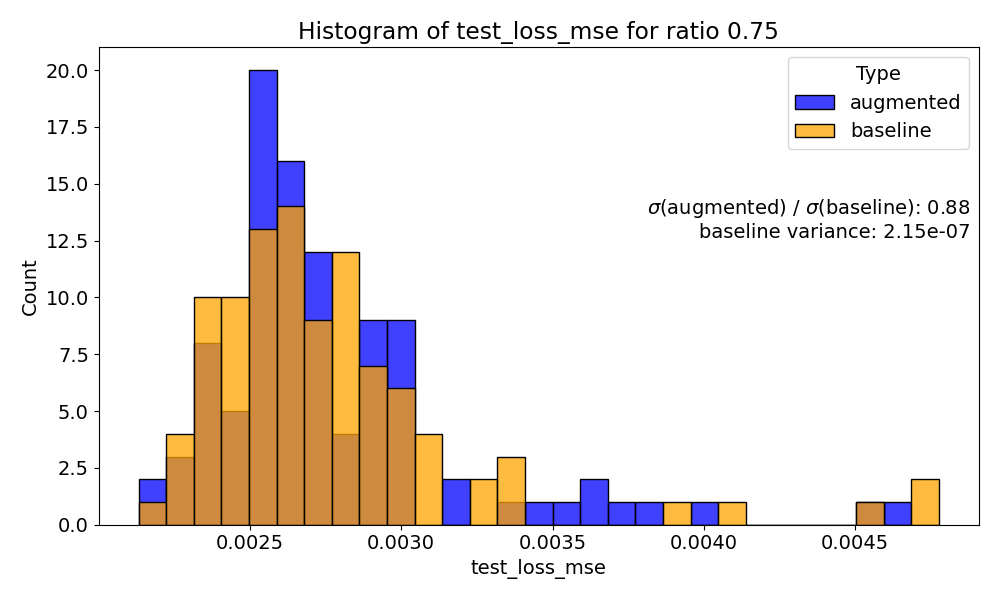}
    \caption{Histogram: Analysis of imbalanced vs augmented forecasting - Ablation ratio of .75 - Results for 100 runs.}
    \label{fig:dist_analysis_exp2_r0.75}
\end{figure}

\input{artifacts/exp2_table_1}

\input{artifacts/exp2_table_2}

%% file: artifacts/exp1_table_mu.tex
\begin{table}[H]
\centering
\caption{Experiment 1 results - aggregated means}
\label{table-mean-exp1}
\begin{tabular}{lccccc}
\toprule
\textbf{Type} & \textbf{test\_mse} & \textbf{test\_mase} & \textbf{test\_mae} & \textbf{test\_mape} \\
\midrule
trtr & 0.003119 & 2.390434 & 0.037563 & 0.251342 \\
tstr & 0.001791 & 2.027180 & 0.030242 & 0.199215 \\
trstr & \textbf{0.001714} & \textbf{1.854287} & \textbf{0.028266} & \textbf{0.162576} \\
\bottomrule
\end{tabular}
\end{table}

%% file: artifacts/exp2_table_var.tex
\begin{table}[H]
\centering
\caption{Experiment 1 results - aggregated standard deviations}
\label{table-var-exp1}
\begin{tabular}{lccccc}
\toprule
\textbf{Type} & \textbf{test\_mse} & \textbf{test\_mase} & \textbf{test\_mae} & \textbf{test\_mape} \\
\midrule
trtr & 0.000579 & \textbf{0.290714} & \textbf{0.003708} & \textbf{0.012571} \\
tstr & \textbf{0.000570} & 0.346175 & 0.005047 & 0.044246 \\
trstr & 0.000756 & 0.392574 & 0.005974 & 0.032716 \\
\bottomrule
\end{tabular}
\end{table}

%% file: artifacts/exp2_table_1.tex
\begin{table}[H]
\centering
\caption{Imbalancing experiment on class 0 results - aggregated means}
\label{table-mean-exp2}
\begin{tabular}{lccccc}
\toprule
\textbf{Type} & \textbf{Ratio} & \textbf{0.25} & \textbf{0.50} & \textbf{0.75} & \textbf{1.00} \\
\midrule
Baseline & test\_mae & \textbf{0.04665} & 0.04671 & 0.03457 & \textbf{0.03140} \\
Baseline & test\_mape & 0.31032& 0.29151 & 0.23888 & \textbf{0.25104}\\
Baseline & test\_mase & \textbf{3.17415} & 3.18327 & 2.20503 & \textbf{2.10880} \\
Baseline & test\_mse & \textbf{0.00425} & 0.00418 & 0.00277& \textbf{0.00217} \\
\midrule
Test run & test\_mae & 0.04677 & \textbf{0.04547} & \textbf{0.03447} & 0.03243 \\
Test run & test\_mape & \textbf{0.31002} & \textbf{0.28877} & \textbf{0.23835} & 0.25472 \\
Test run & test\_mase & 3.17502 & \textbf{3.07901} & \textbf{2.19738} & 2.18728 \\
Test run & test\_mse & 0.00438 & \textbf{0.00399} & 0.00277 & 0.00229 \\
\bottomrule
\end{tabular}

\end{table}

%% file: artifacts/exp2_table_2.tex
\begin{table}[H]
\centering
\caption{Imbalancing experiment on class 0 results - aggreated standard deviations}
\label{table-var-exp2}
\begin{tabular}{lccccc}
\toprule
\textbf{Type} & \textbf{Ratio} & \textbf{0.25} & \textbf{0.50} & \textbf{0.75} & \textbf{1.00} \\
\midrule
Baseline & test\_mae & \textbf{0.01074} & 0.00918 & 0.00463 & 0.00535 \\
Baseline & test\_mape & \textbf{0.03297} & 0.03034 & \textbf{0.02228} & 0.03348 \\
Baseline & test\_mase & \textbf{0.86050} & 0.75173 & 0.36545 & 0.42427 \\
Baseline & test\_mse & \textbf{0.00181} & 0.00137 & 0.00047 & 0.00052 \\
\midrule
Test run & test\_mae & 0.01213 & \textbf{0.00857} & \textbf{0.00431} & \textbf{0.00520 }\\
Test run & test\_mape & 0.03597 & \textbf{0.02768} & 0.02324 & \textbf{0.03209} \\
Test run & test\_mase & 0.94686 & \textbf{0.70215} & \textbf{0.33682} & \textbf{0.42085} \\
Test run & test\_mse & 0.00226 & \textbf{0.00126} & \textbf{0.00044} & \textbf{0.00056} \\
\bottomrule
\end{tabular}
\end{table}

%% file: sections/conclusion.tex
Our experiments highlight the superior performance of a VQVAE-based model compared to some state-of-the-art GAN models for synthesising uni-variate time series in low data environments. Additionally, we discover that in relatively simple datasets, augmenting the dataset with synthetic samples can lead to enhanced forecasting accuracy in subsequent tasks particularly in cases of data scarcity, albeit the expanse of training variance. This latter problem requires further investigation as explained in Section \ref{ssec:exp2}. We also find out that using synthesizers to balance out class distribution neither particularly increases or decreases overall performance. However, it's worth noting the imbalance present in our testing set, which might affect the results. Further investigation and experiments are required on the test set to better understand this behaviour. Notably, we deem necessary to run similar experiments on time series datasets from other domains to ensure validity of our conclusions.

